\documentclass[journal]{IEEEtran}
\usepackage{epsfig}
\usepackage{graphicx}
\usepackage{epstopdf}
\usepackage{multirow}
\usepackage{footmisc}
\usepackage{subfigure}

\ifCLASSINFOpdf
\else
\fi
\begin{document}
%
% paper title
% Titles are generally capitalized except for words such as a, an, and, as,
% at, but, by, for, in, nor, of, on, or, the, to and up, which are usually
% not capitalized unless they are the first or last word of the title.
% Linebreaks \\ can be used within to get better formatting as desired.
% Do not put math or special symbols in the title.
\title{TrajectoryNet: a new spatio-temporal feature learning network for human motion prediction}
%
%
% author names and IEEE memberships
% note positions of commas and nonbreaking spaces ( ~ ) LaTeX will not break
% a structure at a ~ so this keeps an author's name from being broken across
% two lines.
% use \thanks{} to gain access to the first footnote area
% a separate \thanks must be used for each paragraph as LaTeX2e's \thanks
% was not built to handle multiple paragraphs
%

\author{Xiaoli~Liu,~ %~\IEEEmembership{Member,~IEEE,}
        Jianqin~Yin,~ %~\IEEEmembership{Member,~IEEE,}
        Jin~Liu,~%~\IEEEmembership{Member,~IEEE,}
        Pengxiang~Ding,~%~\IEEEmembership{Member,~IEEE,}
        Jun~Liu,~ and ~Huaping~Liu%~\IEEEmembership{Member,~IEEE}% <-this % stops a space
\thanks{
Xiaoli~Liu, Jianqin~Yin,~Jin~Liu and Pengxiang Ding are with the Automation School of Beijing University of Posts and Telecommunications, No.$10$ Xitucheng Road, Haidian District, Beijing $100876$, China.

Jun~Liu is with City University of Hong Kong, Hong Kong $999077$, China.

Huaping Liu is with the Department of Computer Science and Technology of Tsinghua University, Beijing $100084$, China.

e-mail: Liuxiaoli134@bupt.edu.cn, jqyin@bupt.edu.cn. % ,~jinliu@bupt.edu.cn,~dingpx2015@bupt.edu.cn,~Jun.Liu@cityu.edu.hk, hpliu@tsinghua.edu.cn

Jianqin Yin is the corresponding author.}% <-this % stops a space
}

% note the % following the last \IEEEmembership and also \thanks -
% these prevent an unwanted space from occurring between the last author name
% and the end of the author line. i.e., if you had this:
%
% \author{....lastname \thanks{...} \thanks{...} }
%                     ^------------^------------^----Do not want these spaces!
%
% a space would be appended to the last name and could cause every name on that
% line to be shifted left slightly. This is one of those "LaTeX things". For
% instance, "\textbf{A} \textbf{B}" will typeset as "A B" not "AB". To get
% "AB" then you have to do: "\textbf{A}\textbf{B}"
% \thanks is no different in this regard, so shield the last } of each \thanks
% that ends a line with a % and do not let a space in before the next \thanks.
% Spaces after \IEEEmembership other than the last one are OK (and needed) as
% you are supposed to have spaces between the names. For what it is worth,
% this is a minor point as most people would not even notice if the said evil
% space somehow managed to creep in.

% The paper headers
\markboth{Journal of \LaTeX\ Class Files,~Vol.~14, No.~8, January~2020}%
{Shell \MakeLowercase{\textit{et al.}}: Bare Demo of IEEEtran.cls for IEEE Journals}
% The only time the second header will appear is for the odd numbered pages
% after the title page when using the twoside option.
%
% *** Note that you probably will NOT want to include the author's ***
% *** name in the headers of peer review papers.                   ***
% You can use \ifCLASSOPTIONpeerreview for conditional compilation here if
% you desire.

% If you want to put a publisher's ID mark on the page you can do it like
% this:
%\IEEEpubid{0000--0000/00\$00.00~\copyright~2015 IEEE}
% Remember, if you use this you must call \IEEEpubidadjcol in the second
% column for its text to clear the IEEEpubid mark.

% use for special paper notices
%\IEEEspecialpapernotice{(Invited Paper)}

% make the title area
\maketitle

% As a general rule, do not put math, special symbols or citations
% in the abstract or keywords.
\begin{abstract}
   Human motion prediction is an increasingly interesting topic in computer vision and robotics. In this paper, we propose a new $2$D CNN based network, TrajectoryNet, to predict future poses in the trajectory space. Compared with most existing methods, our model focuses on modeling the motion dynamics with coupled spatio-temporal features, local-global spatial features and global temporal co-occurrence features of the previous pose sequence. Specifically, the coupled spatio-temporal features describe the spatial and temporal structure information hidden in the natural human motion sequence, which can be mined by covering the space and time dimensions of the input pose sequence with the convolutional filters. The local-global spatial features that encode different correlations of different joints of the human body (e.g. strong correlations between joints of one limb, weak correlations between joints of different limbs) are captured hierarchically by enlarging the receptive field layer by layer and residual connections from the lower layers to the deeper layers in our proposed convolutional network. And the global temporal co-occurrence features represent the co-occurrence relationship that different subsequences in a complex motion sequence are appeared simultaneously, which can be obtained automatically with our proposed TrajectoryNet by reorganizing the temporal information as the depth dimension of the input tensor. Finally, future poses are approximated based on the captured motion dynamics features. Extensive experiments show that our method achieves state-of-the-art performance on three challenging benchmarks (e.g. Human$3.6$M, G${3}$D, and FNTU), which demonstrates the effectiveness of our proposed method. The code will be available if the paper is accepted.
\end{abstract}

% Note that keywords are not normally used for peerreview papers.
\begin{IEEEkeywords}
human motion prediction, spatio-temporal feature learning, CNN, skeleton.
\end{IEEEkeywords}

% For peer review papers, you can put extra information on the cover
% page as needed:
% \ifCLASSOPTIONpeerreview
% \begin{center} \bfseries EDICS Category: 3-BBND \end{center}
% \fi
%
% For peerreview papers, this IEEEtran command inserts a page break and
% creates the second title. It will be ignored for other modes.
\IEEEpeerreviewmaketitle

\section{Introduction}
\begin{figure}[!t]
\begin{center}
\includegraphics[width=0.9\columnwidth,height=0.9in,trim =50mm 12mm 35mm 75mm, clip=true]{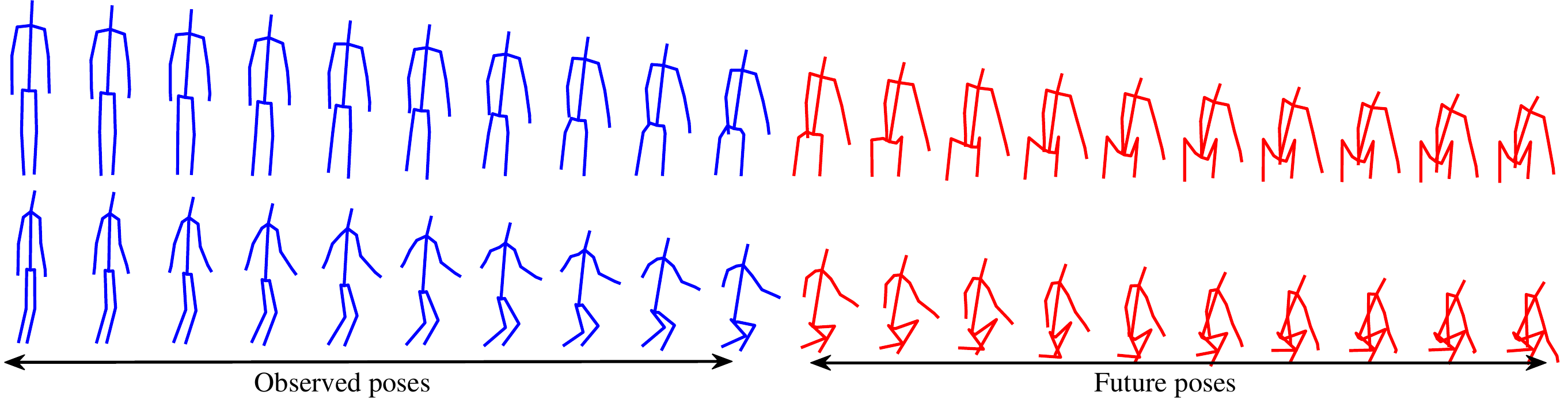}
\end{center}
\caption{Human motion prediction. The blue poses are the observed poses, and the red poses are the future poses.}
\label{fig1a}
\end{figure}

\IEEEPARstart{H}{uman} activity analysis has been an important topic in computer vision due to the undeniable significance in a number of applications, ranging from biomechanics of human movement \cite{bio06}, video surveillance \cite{I3D,twostream14}, to human-machine interaction \cite{APsurvey,hier15}, and service robotics \cite{APsurvey,hier15}. Among many problems in human activity analysis, how to predict future human motion based on the currently observed poses is of great importance to enable automated systems or robots to seamlessly interact with people \cite{RAP18,Liuske19}. For instance, a service robot can provide immediate support to an elderly person to avoid danger if the system is able to predict the person is likely to fall. In this paper, as is shown in Fig. \ref{fig1a}, we focus on the problem of human motion prediction which aims to predict the future human motion sequence based on the observed human motion sequence. 

%Fig. \ref{fig1b} is the workflow of our proposed model, which aims to predict the 3D tensor of the future human motion sequence based on the 3D tensor of the previous human motion sequence.

%\begin{figure}[!t]
%\begin{center}
%\includegraphics[width=0.9\columnwidth,trim = 0mm 0mm 0mm 5mm, clip=true]{figure1_2}
%\end{center}
%\caption{The workflow of our proposed model.}
%\label{fig1b}
%\end{figure}

Coupled spatio-temporal modeling plays a key role to predict future poses \cite{ButepageDRL,pisepp,predcnn}. In general, previous spatio-temporal modeling used two major types of methods, RNN (Recurrent Neural Network) \cite{hmprnn,Fewpr,srnnap,Ntmap,vred} and CNN (Convolutional Neural Network) \cite{ButepageDRL,pisepp,predcnn,cstosap}. ($1$) {\bf RNN models} \cite{hmprnn,Fewpr,Ntmap} are especially powerful in processing short-term temporal information while having an inherent weakness in spatial modeling. For example, Martinez et al. \cite{hmprnn} built their RNN model to predict future poses entirely based on GRU (Gated Recurrent Unit). This method mainly focused on capturing temporal information and ignored a part of the spatial structure of the human body. ($2$) {\bf CNN models} \cite{pisepp,predcnn,cstosap} have been successfully used to predict human poses, but fail to capture the coupled spatio-temporal structure information well. In previous research, Xu et al. \cite{predcnn} and Liu et al. \cite{pisepp,predcnn} proposed CNN based methods and processed the spatial and temporal information separately for pose prediction. However, when a human pose changes across multiple frames, the spatial and temporal information are intrinsically coupled. Separately processing of the spatial and temporal information inevitably breaks the natural state of the information when representing the pose changes, and therefore is ineffective to predict complicated motions.

Another important aspect of human motion prediction is the global temporal co-occurrence modeling. In a complex human motion, some basic motions occur simultaneously in a given temporal context. For example, for a complex motion `` hug'', the basic motions ``grab'' and ``touch'' often happen in the same pose sequence. Because the basic motions may lie in the given full temporal context, we name the co-occurrence relationship of these basic motions as ``global temporal co-occurrence''.
However, most of the existing works model the global temporal information of previous frames in a hierarchical manner \cite{pisepp,predcnn,Tnap}. They modeled different time scales information in different temporal levels and shared weight in each temporal level, which is difficult to learn free parameters for each frame to capture the global temporal co-occurrence relationship of subsequences hidden in the previous motion sequence. 
Besides, Butepage et al. \cite{ButepageDRL} proposed a temporal encoder to model diffident time scales information, which heavily depends on the size of the convolution kernel. However, it has been shown that the smaller the convolution kernel and the deeper the network, the better the final performance of the network \cite{filter16,rethinking16,vgg14}. Thus their model can not capture the global temporal co-occurrence information efficiently.
Some works focused on modeling the spatial co-occurrence information of the human body \cite{CoFL,colstm} and ignored the temporal co-occurrence modeling. For example, Li et al. \cite{CoFL} and Zhu et al. \cite{colstm} proposed to model the spatial co-occurrence information to describe actions. However, this method lacked the modeling of global temporal information and therefore failed to predict pose changes in a long period.

To address the aforementioned limitations, we propose a new spatio-temporal feature learning network, TrajectoryNet, to predict future poses end to end. With this proposed network, the motion dynamics can be captured by encoding the coupled spatio-temporal features, local-global spatial features and global temporal co-occurrence information from a sequence of poses simultaneously. To better mine the motion dynamics hidden in the natural poses sequence, the input pose sequence is represented in the trajectory space. And the spatio-temporal features of previous poses will be extracted with our proposed network in the trajectory space. Specifically, the coupled spatio-temporal information can be captured by covering the coordinates, joints, and time dimensions simultaneously. The local-global spatial features are extracted to model the strong correlations between the joints of the same limb, and the weak correlations between the joints of different limbs. In our new method, the global temporal co-occurrence features can be conveniently captured using CNN by specifying the temporal dimension as channels. Finally, future poses can be predicted using the captured spatio-temporal features by our proposed network.

%The biggest difference from the existing research is that our new proposed network can simultaneously model the coupled spatio-temporal information and global temporal co-occurrence dependency, and also considers the different correlations of joints of the human body.
The main contributions of this research are highlighted as follows:

($1$) A new spatio-temporal learning scheme is proposed to mine the latent motion dynamics in a natural human motion sequence by simultaneously capturing their coupled spatio-temporal structure, different correlations of different joints, and their global temporal co-occurrence relationship of subsequences in the same temporal context.

($2$) A new simple but effective $2$D CNN based network, TrajectoryNet, is proposed to model motion dynamics and predict future poses end to end.

($3$) Experimental results show that our proposed method achieves state-of-the-art performance on three benchmark datasets, which demonstrates the effectiveness of our proposed method.

The remainder of this work is organized as follows: Section II summarises the related literature on human motion prediction. Section III describes the proposed TrajectoryNet to predict future human motion. Section IV reports experimental results on three benchmark datasets both quantitatively and qualitatively. Section V briefly concludes our paper.

\section{Related Work}
In order to accurately predict the human motion, researchers have done significant investigations. We review these works from two aspects: the spatio-temporal modeling and the long-term modeling of human motion.

\subsection{The spatio-temporal modeling}

RNN models have shown their power in short-term temporal modeling and successfully used to predict human motion \cite{hmprnn,Tnap,tprnn,rnmhy,bigan}. Due to the inherent weakness of RNNs, they failed to capture the spatial features of the human body and long-term dependencies efficiently. For example, Chiu et al. \cite{tprnn} modeled the latent hierarchical structure of human motion by capturing the temporal dependencies with different time-scales hierarchically using LSTM cells but failed to capture the spatial structure of the human body. Martinez et al. \cite{hmprnn} proposed a residual architecture to model the velocities of the human motion sequence using GRUs. But the author only focused on short-term modeling and ignored the long-term temporal dependencies and spatial modeling of the human body. To alleviate these limitations hidden in RNNs, some literature is proposed \cite{srnnap,Ntmap,GuiAdversarial,pllsrtd}. Jain et al. \cite{srnnap} proposed a structural-RNN model to model the high-level spatio-temporal structure of the human motion sequence by combining LSTMs and fully connected (FC) layers. Guo et al. \cite{pllsrtd} modeled the local structure of the human body using FC layers and captured the long-term dependencies with GRUs, but failed to capture the interactions between different limbs.

 Many works were proposed to address the spatio-temporal modeling of sequence-to-sequence models using CNN frameworks \cite{ButepageDRL,cstosap,tslap,ostdfvp,mmarnn,d21}. There are two schemes to model the spatio-temporal information: modeling the spatial and temporal information separately, modeling the spatial and temporal information equally. For example, Cho et al. \cite{tslap} proposed to model the spatial and temporal information of previous frames separately. In detail, they first modeled the spatial information using VGG-$16$ \cite{vgg14}, and then proposed T-CNN (Temporal Convolutional Neural Network) to model the temporal dynamics information of previous frames by using convolution across time. Butepage et al. \cite{ButepageDRL} and Li et al. \cite{cstosap} proposed to simultaneously model the spatial and temporal information using CNN and fully connected networks respectively, which may take the spatial and temporal dimension equally and ignore a part of spatio-temporal structure information. Moreover, Mao et al. \cite{TrajD2019} first modeled the temporal information of the human motion sequence using DCT (Discrete Cosine Transform), then captured the spatial dependencies of joint trajectories using GCNs (Graph Convolutional Networks), and achieved state-of-the-art performance. But their model was not constructed end to end, which is not flexible enough to capture the spatio-temporal features of the human motion sequence.

 \subsection{ The long-term modeling of human motion}
 Most of the existing pose prediction models based on RNN (such as LSTM, GRU, i.e.) are inherently hard to capture the long-term dependencies of previous frames using its recurrent unit \cite{hmprnn,Fewpr}.

 Other methods have been reported to model the long-term dynamic information of previous poses hierarchically \cite{pisepp,predcnn,Tnap,tprnn}. Because of the weight sharing mechanism in each temporal level, these models can not learn free parameters for each frame. Therefore, it is difficult to mine the temporal co-occurrences relationships of subsequences from all frames efficiently.
 For example, Xu et al. \cite{predcnn} and Liu et al. \cite{pisepp} proposed to capture the temporal information of adjacent frames using CMU (Cascade Multiplicative Unit) \cite{predcnn}, and model the global temporal information using CMUs hierarchically. In each temporal level, CMUs in their proposed model shared parameters. Therefore, it was difficult to capture the temporal co-occurrence information of all previous frames.

 Moreover, other researchers modeled the global temporal evolution of all history poses \cite{ButepageDRL,mmarnn,ecnnap}. Butepage et al. \cite{ButepageDRL} proposed to capture the temporal information of all input frames using multiple fully connected layers. Li et al. \cite{ecnnap} proposed to model the long-term information of previous frames by mapping the input sequence into deep features using autoencoder \cite{Autoencoder}. Butepage et al. \cite{ButepageDRL} and Li et al. \cite{ecnnap} treated the spatial and temporal information equally, and ignored the difference between spatial and temporal dimension so that their model can not capture the strong long-term temporal dependencies of all input poses efficiently.

 Recently, some researchers modeled temporal information using dilated convolutions \cite{mmarnn}. For example, Pavllo et al. \cite{mmarnn} proposed a QuaterNet to predict future human motion by modeling the long-term temporal information of the input pose sequence hierarchically using dilated convolutions.

 Although great success has been made in long-term modeling, little study has been done to analyze the temporal co-occurrence information of the previous sequence. This motivates us to design a new model that enables the network to get the global response from all input frames to mine the global temporal co-occurrence features of the human motion sequence.

\section{Methodology}

\subsection{Problem formulation}
Human motion sequence can be represented by a group trajectories of a set of $3$D joints. Given an input pose sequence ${S = \{ {p_1},{p_2}, \cdots ,{p_{{N_{t_i}}}}\}}$ with length ${N_{t_i}}$ and its corresponding future pose sequence ${\widehat S = \{ {\widehat p_1},{\widehat p_2}, \cdots ,{\widehat p_{{N_{t_o}}}}\}}$ with length ${N_{t_o}}$, where ${{p_{{i_1}}}{\rm{ = }}\left\{ {{J_{{i_1}{k_1}}}} \right\}_{{k_1} = 1}^{N_j}}$ and ${{\widehat p_{{i_2}}}{\rm{ = }}\left\{ {{{\widehat J}_{{i_2}{k_2}}}} \right\}_{{k_2} = 1}^{N_j}}$ are the ${i_1}$th pose of ${S}$ and ${i_2}$th pose of ${\widehat S}$, respectively, ${{J_{{i_1}{k_1}}} = ({x_{{i_1}{k_1}}},{y_{{i_1}{k_1}}},{z_{{i_1}{k_1}}})}$ is the ${k_1}$th joint of the ${i_1}$th pose of ${S}$, ${{\widehat J_{{i_2}{k_2}}} = ({x_{{i_2}{k_2}}}, {y_{{i_2}{k_2}}}, {z_{{i_2}{k_2}}})}$ is the ${k_2}$th joint of the ${i_2}$th pose of ${\widehat S}$, ${N_j}$ is the number of joints. Then human motion prediction can be formulated as a mapping ${S \to \widehat S}$ from the previous pose sequence $S$ to the future pose sequence ${\widehat S}$.

Most of the existing methods were commonly proposed based on the Encoder-Decoder framework. The encoder was usually used to encode the previous poses into an intermediate representation that represents the spatio-temporal dynamics of previous poses, and the decoder was used to restore the spatial and temporal information of future poses. Following this framework, in this paper, we propose a new network, TrajectoryNet, to model motion dynamics of input pose sequence and predict future motion sequence end to end in the trajectory space. We model the motion dynamics law of human motion by encoding the coupled spatio-temporal information and global temporal co-occurrence features of previous poses and also modeling the different correlations of joints of the human body.

\subsection{Motion dynamics modeling with CNN}
To model the motion dynamics, we extract the spatio-temporal features of the input sequence from three folds:

($1$) {\bf Coupled spatio-temporal features}: as is shown in Fig. \ref{fig1a}, human motion sequence can be considered as the evolution of the pose with time, which contains both spatial and temporal information. Therefore, the human motion sequence is essentially a coupled spatio-temporal structure. Using the convolutional operation, the coupled spatio-temporal features can be easily captured from the width, height, and channels dimension simultaneously from the motion sequence.

 ($2$) {\bf Local-global spatial features}: the joints of the same limb have strong correlations, and the joints of different limbs have weak correlations. For example, the correlation between joints ``left arm'' and ``left wrist'' is stronger than that of joints ``right foot'' and ``left arm''. It is important to model these different correlations of different joints. To model the strong correlations of the joints from the same limb, the joints of the same limb are covered with the convolutional filter, generating features to model the strong correlations of the same limb. Because these features describe the correlations of the same limb, we name them the local spatial features. To model the weak correlations of the joints from different limbs, we gradually enlarge the receptive field of CNN, which can cover the joints from different limbs. The generated features describe the correlations between different limbs, we name them the global spatial features. To model different correlations of the joints, it is necessary to capture the local-global spatial features.

($3$) {\bf Global temporal co-occurrence features}: In this paper, the global temporal co-occurrence features describe the co-occurrence relationship of basic motions of the complex human motion sequence lied in a full temporal context, which can improve the performance of understanding the complex human motion. For a complex motion like ``cooking a meal'', the basic motions ``get'' and ``wash'' may occur in the given temporal context.
These features can be captured automatically using CNN by specifying the temporal dimension of the input sequence as channels. So that the network can get the global response from all frames to exploit the latent temporal co-occurrence relationship of these basic motions in the complex human motion sequence.

\subsection{Architecture of TrajectoryNet}

\begin{figure}[!t]
\begin{center}
\includegraphics[width=\columnwidth,height=1.8in]{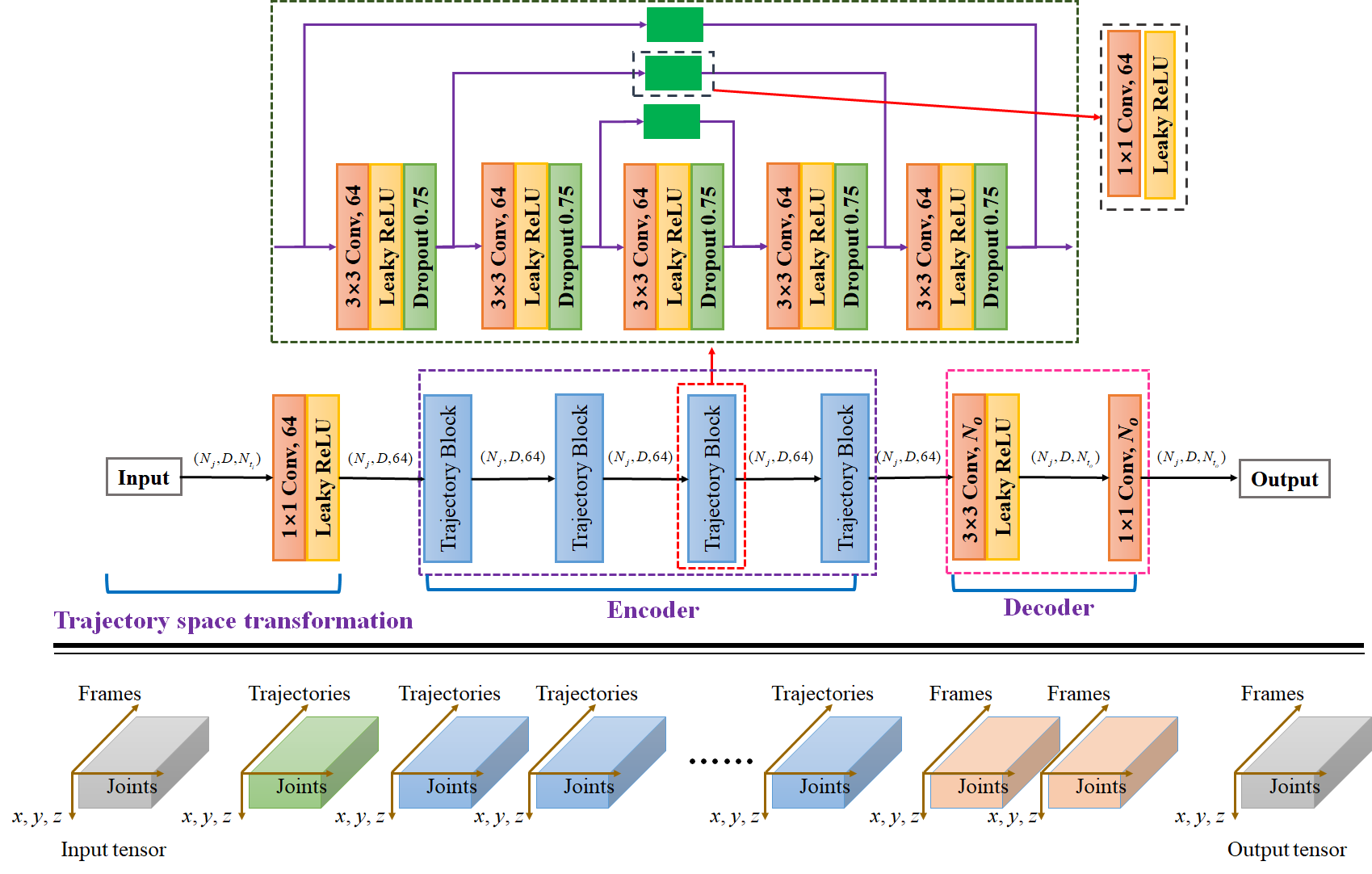}
\end{center}
\caption{Overall Architecture. Top: the architecture of our proposed TrajectoryNet. Bottom: the features map of the corresponding layer. Convolutional layers are in orange, such as ($3 \times 3$ Conv, $64$), where ``$3 \times 3$ Conv'' denotes convolutional operation with $3 \times 3$ filter, and $64$ denotes the output channels. The gray box denotes the input tensor or the output tensor, where the width, height, and channel represent the joints, coordinates (e.g. $x$, $y$, $z$) and frames, respectively.  The green box denotes the output feature map of trajectory space transformation, the blue boxes denote the output feature map of each trajectory block in the Encoder, and the orange boxes denote the output feature map of each layer in the Decoder.  The width, height, and channel of output feature maps in the trajectory space transformation and the Encoder represent the joints, coordinates, and trajectories of the joints, respectively.}
\label{fig3}
\end{figure}

In this section, a new spatio-temporal convolutional network, TrajectoryNet, is proposed to predict future poses as is shown in Fig. \ref{fig3}, which mainly consists of three parts: trajectory space transformation, Encoder, and Decoder.

\begin{figure}[!t]
\begin{center}
\includegraphics[width=0.9\columnwidth,height=1.5in,trim = 0mm 2mm 0mm 2mm, clip=true]{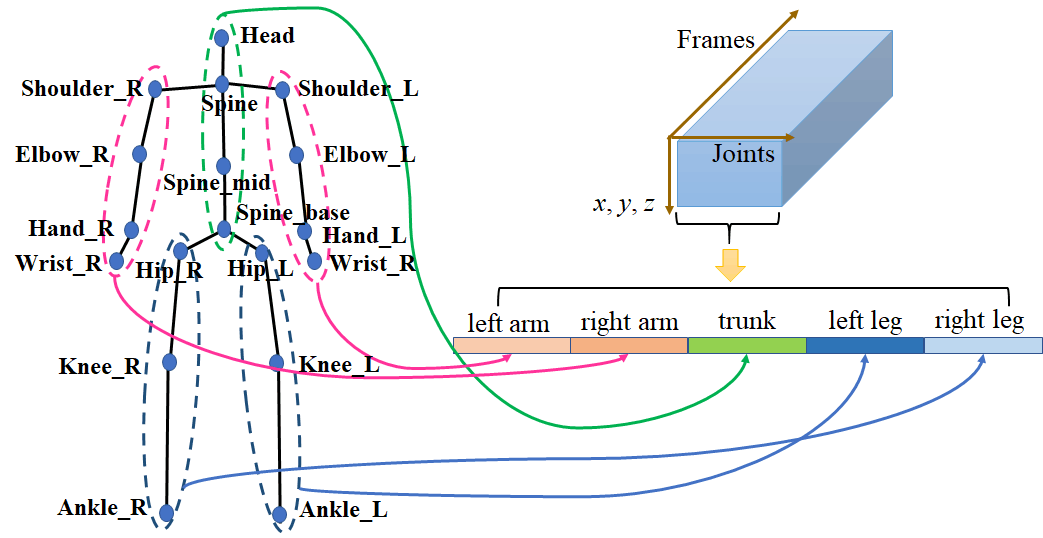}
\end{center}
\caption{Skeletal representation.}
\label{skerepresentation}
\end{figure}

{\bf Skeletal representation:} we first describe the skeletal representation used in our paper.  As is shown in Fig. \ref{skerepresentation}, given a human motion sequence ${S = \{ {p_1},{p_2}, \cdots ,{p_{N_{t_i}}}\}}$ with length ${N_{t_i}}$, these poses can be represented by a tensor ${X = [{p_1};{p_2}; \cdots ;{p_l}; \cdots ;{p_{N_{t_i}}}]}$. Here, as is shown in equation \ref{eqn2}, ${p_l}$ is the ${l}$th pose of the sequence ${S}$. And the shape of the input tensor is ${N_j \times D \times {N_{t_i}}}$. Where ${N_j}$ is the number of joints, ${D}$ is the dimension of each joint. The human body can be naturally divided  into five parts (e.g.  two arms, two legs and trunk) \cite{hier15}. To conveniently model the local-global spatial features of the human body, follow skeletal representation proposed in \cite{pisepp}, as is shown in Fig. \ref{skerepresentation}, joints of the same limb are placed in the connected positions. Finally, the organization order of the human body is: left arm, right arm, trunk, left leg and right leg.

\begin{equation}
{p_l} = \left[ {\begin{array}{*{20}{c}}
{{x_{l,1}}}&{{y_{l,1}}}&{{z_{l,1}}}\\
{{x_{l,2}}}&{{y_{l,2}}}&{{z_{l,2}}}\\
 \vdots & \vdots & \vdots \\
{{x_{l,N_j}}}&{{y_{l,N_j}}}&{{z_{l,N_j}}}
\end{array}} \right],{\rm{ }}l = 1,2, \cdots ,{N_{t_i}}
\label{eqn2}
\end{equation}

\begin{figure}[!t]
\centering
\includegraphics[width=0.9\columnwidth,height=2.1in]{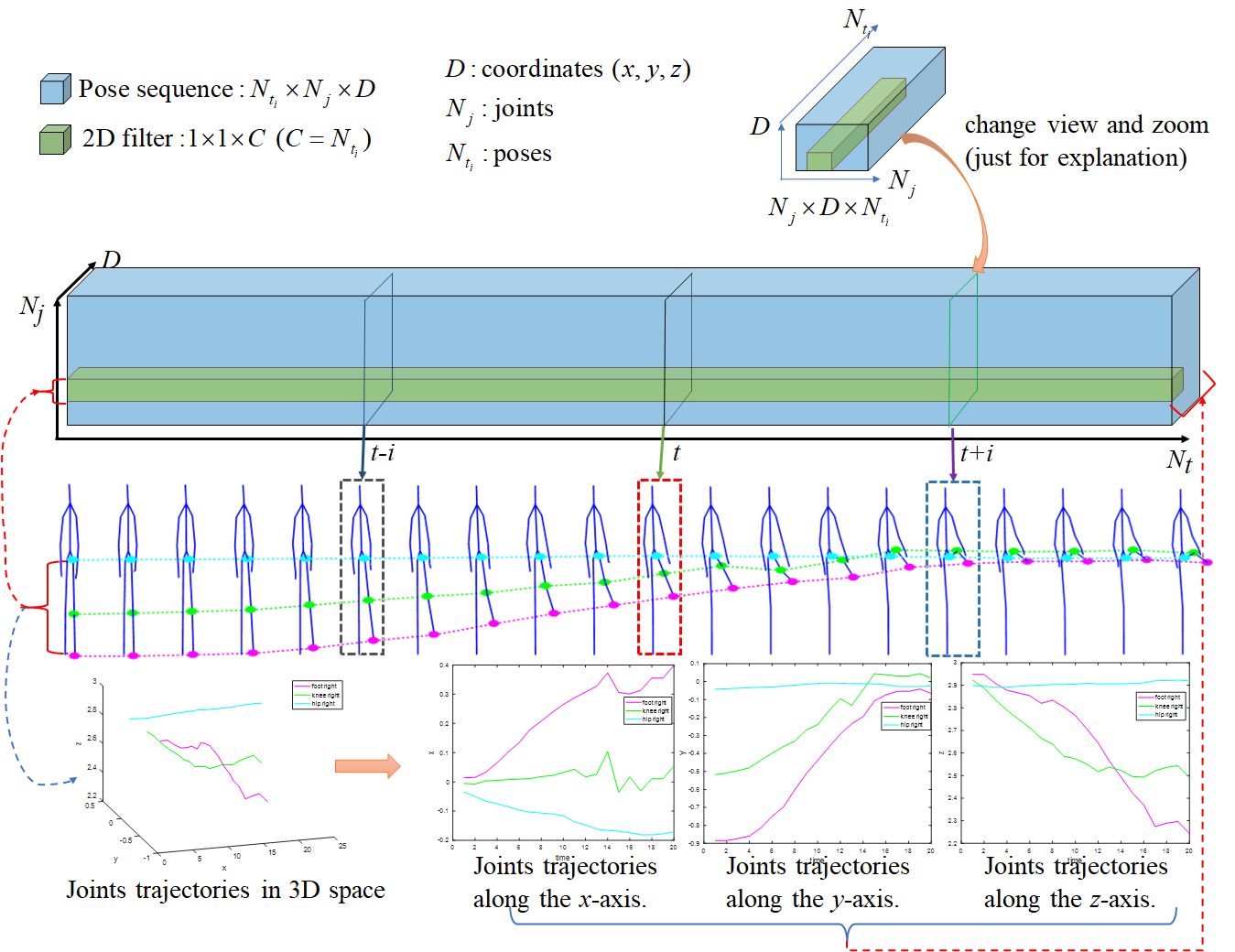}  % width=1.5\columnwidth,height=2.5in
\caption{Trajectory space transformation. Taking the joints of the left leg as an example, from top to bottom, we show a diagram of trajectory space transformation using $2$D convolution, the corresponding human motion sequence, and the joint trajectories in $3$D space and along each axis. Where the filter covers joint trajectories along each axis of each joint across all time-steps. Note: the channel of $2$D filter(i.e. ${C}$) is usually ignored, and we show it explicitly to explain the modeling of joint trajectories.}
\label{fig2}
\end{figure}

{\bf Trajectory space transformation:} in this paper, the trajectory space is defined as follows:

Given a $3$D point $(x,y,z)$ in Euclidean space and its trajectory evolving with time $(x(t),y(t),z(t))$, where $t = 1,2,\cdots, N_t$, $N_t$ is the number of time-steps. Its corresponding point $(x',y',z')$ in the trajectory space that describes the trajectory $(x(t),y(t),z(t))$ information, which can be defined as equation \ref{ts}:

\begin{equation}
(x',y',z') = f(x(t),y(t),z(t))
\label{ts}
\end{equation}
where $f(\cdot)$ is a mapping from the Euclidean space to the trajectory space.

In this trajectory space, each point represents the trajectory information of a special point. Therefore, the trajectory space contains richer trajectory information.
The human motion sequence can be considered as a group of trajectories of joints evolving with time. To better mine their motion dynamics of these joint trajectories, the original motion sequence is converted into the trajectory space. Meanwhile, two findings appear naturally: ($1$) due to the physical constraint of the human body, different joints have different trajectories; ($2$) as is shown in the bottom of Fig. \ref{fig2}, joint trajectories along different axes are different.
Therefore, during space transformation, each joint trajectory and its trajectory along each axis can be treated separately.

In this paper, we make full use of convolutional operation to achieve this space transformation. More specifically, as is shown in Fig. \ref{fig3}, one ${1 \times 1}$ convolutional layer is applied to map the input sequence into the trajectory space. As is shown in Fig. \ref{fig2}, the ${1 \times 1}$ filter only covers the global trajectory of each joint along each axis, which can distinguish: ($1$) one joint trajectory from other joint trajectories; ($2$) one joint trajectory along one axis from other axes. Therefore, each element in the output feature map of this layer encodes the global trajectory information of each joint along each axis. And the output feature maps of this layer encode rich trajectories information of the input pose sequence. As defined in equation \ref{ts}, the transformed information of the input sequence in the trajectory space can be noted as $({x'_{j}},{y'_{j}},{z'_{j}})$, where $j = 1,2,\cdots,N_j$, $(x'_{j},y'_{j},z'_{j})$ is the $j$th joint trajectory information in this trajectory space.

{\bf Encoder:} the Encoder aims to mine the motion dynamics of the human motion sequence in the trajectory space. And the human motion sequence will be encoded into a latent representation $T$, which can be formulated as equation \ref{eqn1}.
\begin{equation}
\label{eqn1}
{T} = \phi ({x'_{j}},{y'_{j}},{z'_{j}}), j = 1,2, \cdots, N_{j}
\end{equation}
Where ${\phi ( \cdot )}$ denotes the Encoder.

As is shown in Fig. \ref{fig3}, our Encoder mainly consists of four trajectory blocks. Each trajectory block mainly consists of five $2$D convolutional layers, and each convolutional layer is followed by a Leaky ReLU and Dropout layer to improve the performance of the network and avoid overfitting.

The motion dynamics can be captured from these perspectives:

($1$) {\bf Global temporal co-occurrence modeling}: after trajectory space transformation, the width, the height and the channel of the output feature map (e.g. the input tensor of the Encoder) represent the joints, coordinates, and trajectories of the joints, respectively. And each point of the input tensor of the Encoder encodes the global trajectory of one joint along an axis. Therefore, with our proposed convolutional Encoder, the global temporal co-occurrence features
can be captured efficiently in the trajectory space.

($2$) {\bf Coupled spatio-temporal modeling}: as is shown in the bottom of Fig. \ref{fig3}, the width, the height and the channel of the feature maps in the Encoder represent the joints, coordinates, and trajectories of the joints. So that the filter covers multiple joint trajectories from both spatial and time dimensions simultaneously. Therefore, the coupled spatio-temporal features of the input sequence can be modeled conveniently and efficiently.

($3$) {\bf Local-global spatial modeling}: we extracted the local-global spatial features from two folds:

$a$) {Symmetrical residual design}. The lower layer usually captures fine-grained features (including local features), while the deeper layer captures coarse-grained features (including global features). In this paper, local features are extracted to describe the strong correlations between the joints of the same limb, and global features are extracted to describe the weak correlations between the joints of different limbs. To capture these correlations, as is shown in Fig.  \ref{fig3}, a symmetric trajectory block is presented to mine the local-global features. Here, the deeper layer is connected with the lower layer, which provides a short connection from the lower layer to the deeper layer, so that the coarse-grained features can be enhanced with fine-grained features.

$b$) {Stacking multiple trajectory blocks}. Multiple trajectory blocks are stacked to enlarge the receptive filed to capture a larger spatial context. In this way, the Encoder encodes the relationship of joint trajectories from local to global. In this paper, the filter size is set to $3$, and the stride is $1$. Therefore, the receptive field of the first layer in the first trajectory block is $3$ (cover $3$ joints), and the receptive field of each layer will be increased by $2$. The receptive field of the first layer in the third trajectory block is $23$, which is large enough to cover all joints of the human body (in this paper, $22$ joints (exclude the unchanged joints and repeated joints) are used on Human$3.6$M, $18$ joints are used on G$3$D and FNTU). Therefore, the global relationship of joint trajectories can be learned after the third trajectory block. And the rest trajectory blocks are stacked to enhance the global relationship of joint trajectories.

Finally, the shape of the output tensor in the Encoder is (${N_j}\times{D}\times{64}$), which denotes the learned motion dynamics of the input sequence. Specifically, each element of the output tensor denotes the spatio-temporal features of a joint trajectory along one axis, which not only encodes the correlations between different dimensions of the same joint but also encodes the correlations between different joints of the same limb and the global information of all joints.

{\bf Decoder:} the Decoder aims to reconstruct the spatial and temporal information of the future pose sequence from the latent representation $T$ learned by the Encoder, which can be formulated as equation \ref{eqn3}.

\begin{equation}
{\widehat p_i} = \psi (T),i = 1,2, \cdots ,{N_{t_o}}
\label{eqn3}
\end{equation}
Where $N_{t_o}$ is the length of the future human motion sequence, ${\widehat p_i}$ is the $i$th future pose, ${\psi ( \cdot )}$ denotes the Decoder.

The correlations between local joints from the same limb are stronger than that of the joints from different limbs. Therefore, as is shown in Fig. \ref{fig3}, one $3 \times 3$ convolutional layer is first applied to approximate the spatio-temporal structure of future poses. In this way, the coordinates of each joint of future poses can be linearly combined by the features of local joints. Finally, one convolutional layer with a ${1 \times 1}$ convolutional filter that covers each joint trajectory along each axis is applied to decouple the correlation between different dimensions of the same joint and the correlation between different joints, which can further recover the spatial details of future poses, smooth the trajectory of future poses and also enhance the predictive performance.

\section{Experiments}
 In this section, we first introduce the datasets used in our experiments and implementation details of our work. Then, we compare our method with state-of-the-arts. Next, we carry out some experiments to analyze the contributions of the proposed method. Next, we show the generalization of our proposed method. Finally, we visualize the predictive performance of our model.
\subsection{Datasets and Implementation Details}

{\bf Datasets:} we use Human3.6M, G${3}$D and FNTU datasets to evaluate the performance of our model. ($1$) Human$3.6$M: Human$3.6$M (H$3.6$M) \cite{h36m} is a commonly used dataset for human motion prediction. The dataset consists of $15$ actions performed by $7$ professional actors. This dataset provides accurate $3$D positions of the human body, and each pose consists of $32$ joints. (${2}$) G${3}$D: G${3}$D \cite{g3d} is a gaming dataset collected with Microsoft Kinect. The dataset consists of ${20}$ actions performed by ${10}$ subjects in a controlled indoor environment. Each people performs several times and each sequence may contain multiple actions. In experiments, we process this dataset following the experimental setting in \cite{pisepp}. (${3}$) FNTU: FNTU \cite{pisepp} is a dataset collected from NTU RGB+D \cite{AmirNTU}, each sequence is clipped from the video in NTU RGB+D. The dataset consists of ${18102}$ sequences, ${12001}$ sequences for training, and the rest for testing. More details can be found at \cite{pisepp}.% \footnote{On H$3.6$M, the coordinate system is defined on the human body, so we call this in first person vision. On G$3$D and FNTU, the coordinate system is defined on the camera, so we call this in third person vision.}

 {\bf Implementation Details:} following the setting and processing of \cite{TrajD2019}, all experiments were carried out in $3$D coordinate space on H$3.6$M, G$3$D, and FNTU. In experiments, all models are implemented by TensorFlow. The channels of the convolutional layer in the trajectory space transformation and Encoder are set to $64$. And the channels of the convolutional layer in Decoder are set to $10$ or $25$ for short-term or long-term prediction respectively (i.e. equal to the number of future poses). Since the joints value can be negative, positive and zero, Leaky ReLU is selected as our activation function. Following \cite{TrajD2019}, we repeat the last frame to align the future poses. To train our model, we use MPJPE (Mean Per Joints Position Error) proposed in \cite{h36m} as our loss as is shown in equation \ref{loss}.
All models are trained with Adam optimizer, and the learning rate is initial with ${0.0001}$. We use MPJPE \cite{h36m} in millimeter on H$3.6$M as our metrics, and use MSE (Mean Squared Error) and MAE (Mean Absolute Error) \cite{pisepp} in meter for per sequence on G$3$D and FNTU \footnote{Note the unit of the converted $3$D coordinates data on H$3.6$M is millimeter, and the unit of the skeletal data on G$3$D and FNTU is meter.}. All experimental settings are consistent with the baselines.

\begin{equation}
l = \frac{1}{{{N_j}*{N_o}}}{\sum\nolimits_{n = 1}^{{N_o}} {\sum\nolimits_{j = 1}^{{N_j}} {\left\| {\widehat {{p}}_{n,j} - {p_{n,j}}} \right\|} ^2}}
\label{loss}
\end{equation}
Where $N_j$ is the number of joints, $N_o$ is the number of future poses, ${p_{n,j}}$ is the groundtruth one, and ${\widehat p}_{n,j}$ is the predictive one.

\subsection{Comparison with state-of-the-arts}

{\bf Results on H$3.6$M:} Table \ref{table7} reports the short-term prediction errors for $15$ activities and their average performance. Specifically, compared with the results of the RNN model \cite{hmprnn} and the CNN model \cite{cstosap}, the error of our proposed method is significantly decreased, which demonstrates the effectiveness of our proposed method. These possible reasons are ($1$) RNN model \cite{hmprnn} ignores some spatial information, different correlations of different joints and long-term temporal information; ($2$) CNN model \cite{cstosap} ignores the global temporal co-occurrence modeling; ($3$)But our method captures the motion dynamics of the human motion sequence by learning the coupled spatio-temporal features, local-global spatial features and global temporal co-occurrence features simultaneously. So that our model can capture the motion dynamics law better.
Compared with \cite{TrajD2019}, our model achieves the lowest errors at each time-step. Our model captures the motion dynamics and predicts future poses end to end, while \cite{TrajD2019} captures the motion dynamics and predict future poses separately. Thus our model is more flexible. This may be a possible reason why our model achieves superior performance.

For long-term prediction, as is shown in Table \ref{table11}, compared with all the baselines, our model achieves the best performance on average for both $560$ms and $1000$ms, which further verify the effectiveness of our proposed method.

\begin{table*}[!t]
\caption{Short-term prediction on H$3.6$M.}
\footnotesize
%\label{table4}
%\centering  % {ccccc|cccc|cccc|cccc}
\begin{center}
\begin{tabular}{p{2.8cm}p{0.4cm}p{0.4cm}p{0.4cm}p{0.5cm}|p{0.4cm}p{0.4cm}p{0.4cm}p{0.5cm}|p{0.4cm}p{0.4cm}p{0.4cm}p{0.5cm}|p{0.4cm}p{0.4cm}p{0.4cm}p{0.5cm}}
\hline
\multirow{2}{*}{Milliseconds}& \multicolumn{4}{c}{Walking} & \multicolumn{4}{c}{Eating}& \multicolumn{4}{c}{Smoking} & \multicolumn{4}{c}{Discussion}\\
%\hline
\cline{2-17} & 80 &160 & 320 &400 & 80 &160 & 320 &400& 80 &160 & 320 &400& 80 &160 & 320 &400\\
\hline
Residual sup \cite{hmprnn} &23.8 &40.4& 62.9& 70.9& 17.6& 34.7& 71.9& 87.7& 19.7& 36.6& 61.8& 73.9& 31.7& 61.3& 96.0& 103.5 \\
convSeq2Seq \cite{cstosap} &17.1 &31.2&53.8&61.5&13.7&25.9&52.5&63.3&11.1&21.0&33.4&38.3&18.9&39.3&67.7&75.7 \\
%\hline
LearnTrajDep \cite{TrajD2019}&8.9 &15.7&{\bf 29.2}&{\bf 33.4}& 8.8& 18.9& 39.4& 47.2& 7.8& 14.9& 25.3& 28.7& 9.8& 22.1&{\bf 39.6} &{\bf 44.1} \\
%\hline
TrajectoryNet (Ours)&{\bf8.2}&{\bf 14.9}&30.0&35.4&{\bf 8.5}&{\bf 18.4}&{\bf 37.0 }&{\bf 44.8}&{\bf 6.3}&{\bf 12.8}&{\bf 23.7}&{\bf 27.8}&{\bf 7.5}&{\bf 20.0}&41.3&47.8\\
\hline
\hline
\multirow{2}{*}{Milliseconds}& \multicolumn{4}{c}{Directions} & \multicolumn{4}{c}{Greeting}& \multicolumn{4}{c}{Phoning} & \multicolumn{4}{c}{Posing}\\
%\hline
 \cline{2-17}& 80 &160 & 320 &400 & 80 &160 & 320 &400& 80 &160 & 320 &400& 80 &160 & 320 &400\\
\hline
Residual sup \cite{hmprnn} & 36.5 &56.4& 81.5& 97.3& 37.9& 74.1& 139.0& 158.8 &25.6& 44.4& 74.0& 84.2& 27.9& 54.7& 131.3& 160.8 \\
convSeq2Seq \cite{cstosap} & 22.0&37.2 &59.6& 73.4 &24.5 &46.2 &90.0& 103.1& 17.2& 29.7& 53.4 &61.3& 16.1& 35.6& 86.2& 105.6\\
%\hline
LearnTrajDep \cite{TrajD2019}& 12.6 & 24.4&{\bf 48.2}&{\bf 58.4}& 14.5& 30.5& 74.2& 89.0& 11.5& 20.2& 37.9& 43.2& 9.4& 23.9& 66.2& 82.9\\
%\hline
TrajectoryNet (Ours)&{\bf 9.7}&{\bf 22.3}&50.2&61.7&{\bf 12.6}&{\bf 28.1}&{\bf 67.3}&{\bf 80.1}&{\bf 10.7}&{\bf 18.8}&{\bf 37.0}&{\bf 43.1}&{\bf 6.9}&{\bf 21.3}&{\bf 62.9}&{\bf 78.8}\\
\hline
\hline
\multirow{2}{*}{Milliseconds}& \multicolumn{4}{c}{Purchases} & \multicolumn{4}{c}{Sitting}& \multicolumn{4}{c}{Sitting Down} & \multicolumn{4}{c}{Taking Photo}\\
%\hline
\cline{2-17} & 80 &160 & 320 &400 & 80 &160 & 320 &400& 80 &160 & 320 &400& 80 &160 & 320 &400\\
\hline
Residual sup \cite{hmprnn} & 40.8& 71.8& 104.2& 109.8 &34.5& 69.9& 126.3 &141.6& 28.6& 55.3& 101.6& 118.9 &23.6 &47.4& 94.0& 112.7\\
convSeq2Seq \cite{cstosap} & 29.4& 54.9& 82.2& 93.0 &19.8 &42.4& 77.0& 88.4 & 17.1& 34.9& 66.3& 77.7& 14.0& 27.2& 53.8& 66.2\\
%\hline
LearnTrajDep \cite{TrajD2019}& 19.6& 38.5& 64.4&{\bf 72.2}& 10.7& 24.6& 50.6&{\bf 62.0}&11.4 &{\bf 27.6}& 56.4& 67.6& 6.8& 15.2& 38.2& 49.6\\
%\hline
TrajectoryNet (Ours)&{\bf 17.1}&{\bf 36.1}&{\bf 64.3}&75.1&{\bf 9.0}&{\bf 22.0}&{\bf 49.4}&62.6&{\bf 10.7}&28.8&{\bf 55.1}&{\bf 62.9}&{\bf 5.4}&{\bf 13.4}&{\bf 36.2}&{\bf 47.0}\\
\hline
\hline
\multirow{2}{*}{Milliseconds}& \multicolumn{4}{c}{Waiting} & \multicolumn{4}{c}{Walking Dog}& \multicolumn{4}{c}{Walking Together} & \multicolumn{4}{c}{Average}\\
%\hline
 \cline{2-17}& 80 &160 & 320 &400 & 80 &160 & 320 &400& 80 &160 & 320 &400& 80 &160 & 320 &400\\
\hline
Residual sup \cite{hmprnn} & 29.5& 60.5& 119.9& 140.6& 60.5& 101.9& 160.8& 188.3& 23.5& 45.0& 71.3& 82.8& 30.8& 57.0& 99.8& 115.5\\
convSeq2Seq \cite{cstosap} &17.9& 36.5& 74.9& 90.7& 40.6& 74.7& 116.6& 138.7& 15.0& 29.9& 54.3& 65.8& 19.6& 37.8& 68.1& 80.2\\
%\hline
LearnTrajDep \cite{TrajD2019}& 9.5& 22.0& 57.5& 73.9& 32.2& 58.0& 102.2& 122.7& 8.9& {\bf 18.4}& 35.3& 44.3& 12.1& 25.0& 51.0& 61.3\\
%\hline
TrajectoryNet (Ours)&{\bf 8.2}&{\bf 21.0}&{\bf 53.4}&{\bf 68.9}&{\bf 23.6}&{\bf 52.0}&{\bf 98.1}&{\bf 116.9}&{\bf 8.5}&18.5&{\bf 33.9}&{\bf 43.4}&{\bf 10.2}&{\bf 23.2}&{\bf 49.3}&{\bf 59.7}\\
\hline
\end{tabular}
\end{center}
\label{table7}
\end{table*}

% long-term prediction over 15 activities
\begin{table*}[!t]
\caption{Long-term prediction over $15$ activities on H$3.6$M. The $3$D errors for $4$ activities of ``Residual sup'' and ``convSeq2Seq '' models reported in \cite{TrajD2019} are provided. Moreover, the reported results in \cite{TrajD2019} have shown the performance of \cite{TrajD2019} outperformed that of \cite{hmprnn} and \cite{cstosap} by a large margin. Therefore, we just reproduce the results of \cite{TrajD2019} for all activities using available code.}

\footnotesize
%\label{table4}
%\centering  % {ccc|cc|cc|cc|cc|cc|cc|cc}
\begin{center}
\begin{tabular}{p{2.75cm}p{0.3cm}p{0.5cm}|p{0.3cm}p{0.5cm}|p{0.3cm}p{0.3cm}|p{0.3cm}p{0.3cm}|p{0.3cm}p{0.5cm}|p{0.3cm}p{0.3cm}|p{0.5cm}p{0.5cm}|p{0.3cm}p{0.5cm}}
\hline
\multirow{2}{*}{Milliseconds}& \multicolumn{2}{c}{Walking} & \multicolumn{2}{c}{Eating}& \multicolumn{2}{c}{Smoking} & \multicolumn{2}{c}{Discussion} &\multicolumn{2}{c}{Directions} & \multicolumn{2}{c}{Greeting}& \multicolumn{2}{c}{Phoning} & \multicolumn{2}{c}{Posing}\\
%\hline
\cline{2-17} & 560 &1000 & 560 &1000& 560 &1000& 560 &1000& 560 &1000 & 560 &1000& 560 &1000& 560 &1000\\
\hline
Residual sup \cite{hmprnn} &73.8 &86.7& 101.3& 119.7& 85.0 &118.5& 120.7& 147.6&{--}&{--}&{--}&{--}&{--}&{--}&{--}&{--}\\
convSeq2Seq \cite{cstosap} & 59.2& 71.3& 66.5& 85.4& 42.0& 67.9& 84.1& 116.9&{--}&{--}&{--}&{--}&{--}&{--}&{--}&{--}\\
LearnTrajDep \cite{TrajD2019}& 42.2&51.6&{\bf 57.1}&{\bf 69.5}&{\bf 32.5}&60.7&{\bf 70.5}&{\bf 99.6}& {\bf 79.6}&{\bf 102.9}&95.8& 89.9&62.6&113.8& {\bf 107.2}& 211.8 \\
%\hline
TrajectoryNet (Ours)&{\bf 37.9} &{\bf 46.4} &59.2 &71.5 &32.7 &{\bf 58.7} &75.4 &103 &84.7 &104.2 &{\bf 91.4} &{\bf 84.3} &{\bf 62.3} &{\bf 113.5}&111.6&{\bf 210.9}\\
\hline
\hline
\multirow{2}{*}{Milliseconds}& \multicolumn{2}{c}{Purchases} & \multicolumn{2}{c}{Sitting}& \multicolumn{2}{c}{Sitting Down} & \multicolumn{2}{c}{Taking Photo}& \multicolumn{2}{c}{Waiting} & \multicolumn{2}{c}{Walking Dog}& \multicolumn{2}{c}{Walking Together} & \multicolumn{2}{c}{Average}\\
%\hline
\cline{2-17} & 560 &1000 & 560 &1000& 560 &1000& 560 &1000& 560 &1000 & 560 &1000& 560 &1000& 560 &1000\\
\hline
%\hline
Residual sup \cite{hmprnn} &{--}&{--}&{--}&{--}&{--}&{--}&{--}&{--}&{--}&{--}&{--}&{--}&{--}&{--}&{--}&{--}\\
convSeq2Seq \cite{cstosap} &{--}&{--}&{--}&{--}&{--}&{--}&{--}&{--}&{--}&{--}&{--}&{--}&{--}&{--}&{--}&{--}\\
LearnTrajDep \cite{TrajD2019}& 92.4&125.3&{\bf 78.6}&{\bf 109.9}& 88.1& 137.8& 78.3& 95.0& 99& 169.5& {\bf 139.2}&{\bf 167.7}& 60.3& 84.1& 78.9& 112.6\\
%\hline
TrajectoryNet (Ours) &{\bf 84.5}&{\bf 115.5}&81&116.3&{\bf 79.8}&{\bf 123.8}& {\bf 73.0}&{\bf 86.6}&{\bf 92.9}&{\bf 165.9}& 141.1& 181.3&{\bf 57.6}&{\bf 77.3}&{\bf 77.7}&{\bf 110.6}\\
\hline
\end{tabular}
\end{center}
\label{table11}
\end{table*}

% frame-wise的实验结果可以补充（好像不可以呢）
{\bf Results on G$3$D and FNTU:} as is shown in Table \ref{table1}, our proposed method achieves state-of-the-art performance on both datasets, especially on a larger dataset (e.g. FNTU). Compared with PredCNN$'$ \cite{predcnn,pisepp} and PISEP$^2$ \cite{pisepp}, the error of our method decreases greatly. The possible reasons are two folds: ($1$) Our model captures the spatial and temporal information simultaneously, which considers the spatio-temporal structure carefully. While PredCNN$'$ and PISEP$^2$ separately model the spatial and temporal features of previous poses, which ignores some correlations between spatial features and temporal features. ($2$) Our method models the strong global temporal co-occurrence features of previous poses conveniently using CNN in the trajectory space. While PredCNN$'$ and PISEP$^2$ modeled the global temporal information hierarchically, which is difficult to capture the strong global temporal co-occurrence features of the input sequence.  Compared with \cite{TrajD2019}, our method achieves the best performance on both datasets and yields a large margin on the larger dataset, which demonstrates the effectiveness of our proposed method again.
\begin{table}[!t]
\caption{Human motion prediction on G$3$D and FNTU}
\footnotesize
%\centering
\begin{center}
\begin{tabular}{ccc|cc}
\hline
\multirow{2}{*}{Model}& \multicolumn{2}{c}{G$3$D} & \multicolumn{2}{c}{FNTU} \\
%\hline
 \cline{2-5}& MSE &MAE & MSE &MAE \\
\hline
 PredCNN$'$ \cite{predcnn,pisepp} &0.1882&1.5713&0.1665&1.6394 \\
%\hline
%\hline
 PISEP$^2$ \cite{pisepp}&0.1199&1.1101 &0.1210&1.1651\\
 LearnTrajDep \cite{TrajD2019} &0.1013&0.9068&0.1696&1.2280 \\
TrajectoryNet (Ours)&{\bf 0.0937}&{\bf 0.8663}&{\bf 0.1055}&{\bf 1.0114}\\
\hline
\end{tabular}
\end{center}

\label{table1}
\end{table}

\subsection{Ablation analysis}
In this section, we verify our network from coupled spatio-temporal modeling, global temporal co-occurrence modeling, and symmetrical residual design.

{\bf Evaluation of coupled spatio-temporal modeling:} to verify the importance of coupled spatio-temporal modeling, the filter size is set to $1$ in our network. In this case, the network only focuses on modeling temporal information, while ignores spatial modeling. As is shown in Table \ref{table9}, compared with the results of ``RS'' (where ``RS'' denotes ``remove spatial modeling''), the errors of ``WS'' (where ``WS'' denotes ``with spatial modeling'') decrease significantly on three datasets, which demonstrates that simultaneously modeling the spatial and temporal features is critical for the network performance. For example, ``WS'' on H$3.6$M achieves lower error at all timestamps, especially for the later prediction. The possible reason is: at early timestamps, the future spatial features are similar to the later observed poses, but at later timestamps, the future spatial features may vary greatly. ``RS'' ignores the modeling of spatial information, which may lead to the worse performance at later timestamps.

\begin{table}[!t]
\caption{Evaluation of coupled spatio-temporal modeling. Here, ``RS'' denotes ``remove spatial modeling'',``WS'' denotes ``with spatial modeling''. The errors on H$3.6$M are averaged over $15$ activities.}
\centering
\begin{tabular}{ccccc}
\hline
\multirow{2}{*}{Model}& \multicolumn{2}{c}{G$3$D} & \multicolumn{2}{c}{FNTU} \\
%\hline
 \cline{2-5}& MSE &MAE & MSE &MAE \\
\hline
RS &0.1542&1.0681 &0.1566& 1.2057\\
%\hline
WS &{\bf 0.0937}&{\bf 0.8663}&{\bf 0.1055}&{\bf 1.0114}\\
\hline
\hline

\multirow{2}{*}{Model}& \multicolumn{4}{c}{MPJPE on H$3.6$M} \\
%\hline
 \cline{2-5}& 80 &160 & 320 &400 \\
 \hline
 RS &12.4 &31.0 &67.6 & 81.6 \\
%\hline
WS &{\bf 10.2}&{\bf 23.2}&{\bf 49.3}&{\bf 59.7}\\
\hline
\end{tabular}
\label{table9}
\end{table}

{\bf Evaluation of global temporal co-occurrence modeling:} to show the effectiveness of global temporal co-occurrence information, we reorganize the input tensor (frames are set as width, joints are set as height, and coordinates (e.g. $x$, $y$ and $z$) are set as depth) as done in prior literature\cite{duske,kenew17,liske17}. In this case, the filter covers local spatial and local temporal information, so that we can remove the global temporal co-occurrence modeling of our network. But the receptive field of the network can be increased layer by layer. Therefore, to avoid the influence of depth of the network and carefully analyze the effectiveness of global temporal co-occurrence modeling, we carry out two group experiments: ($1$) ``remove global temporal co-occurrence modeling and use $1$ trajectory block'' (RGCOT-$1$) and ``with global temporal co-occurrence modeling and use $1$ trajectory block'' (WGCOT-$1$); ($2$) ``remove global temporal co-occurrence modeling and use $4$ trajectory blocks'' (RGCOT-$4$) and ``with global temporal co-occurrence modeling and use $4$ trajectory blocks'' (WGCOT-$4$).

Experimental results are reported in Table \ref{table4}. Compared with ``RGCOT-$1$'', the errors of ``WGCOT-$1$'' are decreased significantly on two datasets, which shows the effectiveness of global temporal co-occurrence modeling. However, when the depth of the network is deepened (such as using 4 trajectory blocks), the improvement of global temporal co-occurrence modeling is reduced. For example,  in general, compared with ``RGCOT-$4$'', the performance of ``WGCOT-$4$'' is slightly better on both datasets, but the improvement is limit especially on a smaller dataset (such as FNTU). The possible reasons are: ($1$) when using $4$ trajectory blocks in our proposed TrajectoryNet, the receptive field is enlarged layer by layer. In the last convolutional layer of the first trajectory block, the receptive field is $11$, which is large enough to capture the global temporal information of the previous poses (in all experiments, the length of the input pose sequence is set to $10$). The rest three trajectory blocks are stacked to augment the global temporal information and capture the temporal co-occurrence features of the inputs. ($2$) Compared with a larger dataset (such as H$3.6$m), the prediction task on a smaller dataset (such as FNTU) will be relatively simple. These may be the reasons for the limited performance improvement of global temporal co-occurrence modeling on a small dataset. To sum up, we can conclude that our proposed ``global temporal co-occurrence modeling'' can promote the final performance of the network, especially in a shallow network.

\begin{table}[!t]
\caption{Evaluation of global temporal co-occurrence modeling. Here, ``RGCOT-$1$'' denotes ``remove global temporal co-occurrence modeling and use $1$ trajectory block'', ``RGCOT-$4$'' denotes ``remove global temporal co-occurrence modeling and use $4$ trajectory blocks'', ``WGCOT-$1$'' denotes ``with global temporal co-occurrence modeling and use $1$ trajectory block'', ``WGCOT-$4$'' denotes ``with global temporal co-occurrence modeling and use $4$ trajectory blocks'', $F_i$ denotes $i$th frame. The average error on FNTU are averaged over $10$ predictive frames. The errors on H$3.6$M are averaged over $15$ activities.}
\centering

\begin{tabular}{cccccc}
\hline
\multirow{2}{*}{Model}& \multicolumn{5}{c}{MSE on FNTU} \\
%\hline
 \cline{2-6}& $F_2$ &$F_4$ & $F_8$ &$F_{10}$ &average \\
\hline
 RGCOT-$1$ &0.0903& 0.0897& 0.2002& 0.2877&0.1518\\
%\hline
 WGCOT-$1$ &{\bf 0.0420}&{\bf 0.0759}&{\bf 0.1809}&{\bf 0.2474} & {\bf 0.1230}\\
 \hline
\hline
 RGCOT-$4$ &0.0379 & {\bf 0.0670}& 0.1525& {\bf 0.2092}&{\bf 0.1053}\\
%\hline
 WGCOT-$4$ &{\bf 0.0365}&{ 0.0673}&{\bf 0.1524}&{ 0.2101} &{0.1055}\\
\hline
\hline
\multirow{2}{*}{Model}& \multicolumn{5}{c}{MAE on FNTU} \\
%\hline
 \cline{2-6}& $F_2$ &$F_4$ & $F_8$ &$F_{10}$ &average \\
\hline
 RGCOT-$1$ &{0.9797}&{ 0.9904}&{ 1.5420}&{ 1.8620} & {1.2850}\\
%\hline
 WGCOT-$1$ &{\bf 0.5631}&{\bf 0.8722}&{\bf 1.4592}&{\bf 1.7370} & {\bf 1.0922}\\
 \hline
\hline
 RGCOT-$4$ &0.5670 & 0.8250& 1.3476& 1.6010& 1.0258\\
%\hline
 WGCOT-$4$ &{\bf 0.5350}&{\bf 0.8160}&{\bf 1.3389}&{\bf 1.5943} & {\bf 1.0114}\\
\hline
\end{tabular}

\begin{tabular}{ccccc}
\\
\hline
\multirow{2}{*}{Model}& \multicolumn{4}{c}{MPJPE on H$3.6$M} \\
%\hline
 \cline{2-5}& 80 &160 & 320 &400 \\
 \hline
 RGCOT-$1$ &33.0 &31.8 & 56.9& 68.3 \\
%\hline
WGCOT-$1$ &{\bf 11.1}&{\bf 25.3}&{\bf 53.4}&{\bf 64.4}\\
 \hline
 RGCOT-$4$ &12.0 &25.5 & 51.1& 60.7 \\
%\hline
WGCOT-$4$ &{\bf 10.2}&{\bf 23.2}&{\bf 49.3}&{\bf 59.7}\\

\hline
\end{tabular}
\label{table4}
\end{table}

\begin{figure}[!t]
\begin{center}
\includegraphics[width=0.8\columnwidth,height=0.5in, trim = 10mm 10mm 8mm 10mm, clip=true]{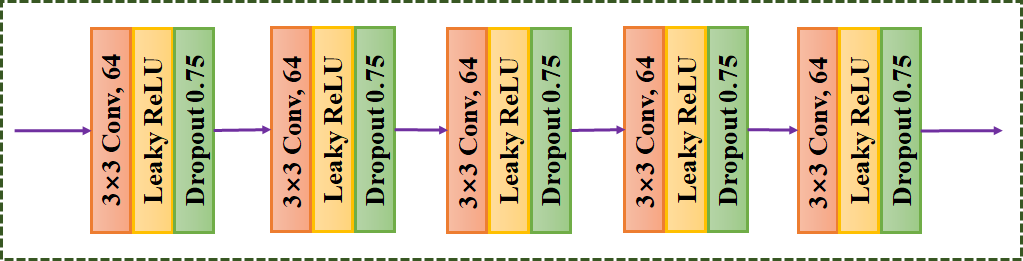}
\end{center}
\caption{Remove residual connections in the trajectory block.}
\label{no_sym}
\end{figure}

{\bf Evaluation of symmetrical residual design:} to verify the effectiveness of our proposed symmetrical residual design in trajectory block, as is shown in Figure \ref{no_sym}, we remove all residual connections in the trajectory block. Experimental results are shown in Table \ref{sym}. Compared with ``RSRC'' (where ``RSRC'' denotes ``remove symmetrical residual connections'' in the trajectory block), the errors of ``WSRC'' (where ``WSRC'' denotes ``with symmetrical residual connections'' in the trajectory block) are significantly decreased on two datasets, which demonstrates the effectiveness of residual connected design in our proposed network.

\begin{table}[!t]
\caption{Evaluation of symmetrical residual design. Here, $F_i$ denotes the $i$th predictive pose, ``RSRC'' denotes ``remove symmetrical residual connections'' in the trajectory block as is shown in Figure \ref{no_sym}, ``WSRC'' denotes ``with symmetrical residual connections'' in the trajectory block. The average errors on FNTU are averaged over all predictive poses (i.e. $10$ future poses). The errors on H$3.6$M are averaged over $15$ activities.}
\centering
\begin{tabular}{p{1.2cm}p{0.8cm}p{0.8cm}p{0.8cm}p{0.8cm}p{1cm}}
\hline
\multirow{2}{*}{Model}& \multicolumn{5}{c}{MSE on FNTU} \\
%\hline
 \cline{2-6}& {$F_2$}& {$F_4$}&{$F_8$}&{$F_{10}$}& average \\
\hline
RSRC& 0.1671&0.1412&0.3373&0.3636&0.2289\\
WSRC&{\bf 0.0365}&{\bf 0.0673}&{\bf 0.1524}&{\bf 0.2101}&{\bf 0.1055}\\
\hline
\hline
\multirow{2}{*}{Model}& \multicolumn{5}{c}{MAE on FNTU} \\
%\hline
 \cline{2-6}& {$F_2$}& {$F_4$}&{$F_8$}&{$F_{10}$}& average \\
\hline
RSRC &1.5053 &1.6507&2.2590&2.4883& 1.9175\\
WSRC &{\bf 0.5350}&{\bf 0.8160}&{\bf 1.3389}&{\bf 1.5943}&{\bf 1.0114}\\
\hline
\end{tabular}

\begin{tabular}{ccccc}
\\
\hline
\multirow{2}{*}{Model}& \multicolumn{4}{c}{MPJPE on H$3.6$M} \\
%\hline
 \cline{2-5}& 80ms &160ms & 320ms &400ms \\
 \hline
RSRC &35.7&37.2 &65.8&79.8\\
WSRC &{\bf 10.2}&{\bf 23.2}& {\bf 49.3}&{\bf 59.7}\\
\hline
\end{tabular}
\label{sym}
\end{table}

\subsection{Evaluation of generalization}  % \footnote{ The coordinates data on G$3$D and FNTU are in the third person vision, while H$3.6$M is in the first vision. Therefore, we only use G$3$D and FNTU to evaluate the generalization performance of our network.}  \footnote{H$3.6$M is a dataset in first person vision, while G$3$D and FNTU are the dataset in third person vision.}.
\begin{table}[!t]
\caption{Evaluation of generalization. In the experiment, we train on FNTU dataset and directly test on G$3$D dataset, which is consistent with the baselines \cite{predcnn,pisepp,TrajD2019}.}
\centering
\begin{tabular}{ccc}
\hline
Methods &MSE&MAE\\
\hline
PredCNN$'$ \cite{predcnn,pisepp} &0.2432&2.1706 \\
%\hline
PISEP$^2$ \cite{pisepp}&0.1446 &1.2713\\
LearnTrajDep \cite{TrajD2019}&0.1353 &1.0409\\
TrajectoryNet (Ours)&{\bf 0.1179 }&{\bf 0.9353 } \\
\hline
\end{tabular}
\label{table6}
\end{table}

Considering the differences of these datasets, we decide to verify the generalization performance of the proposed network on G$3$D and FNTU. For this, we pre-train our model on FNTU and directly test on G$3$D. As is shown in Table \ref{table6}, our method achieves the best performance on unseen data. For example, compared with \cite{TrajD2019}, the MSE and MAE of our network decrease by $0.0174$ and $0.1056$ respectively. The possible reason is that our model considers modeling motion dynamics and predicting future poses end to end, which may lead to better generalization of our proposed model to novel actions.

\subsection{Qualitative analysis}
% 可视化的结果图
\begin{figure*}[!t]
\begin{center}
\subfigure[Greeting]{\includegraphics[width=0.87\columnwidth,height=1in, trim = 50mm 18mm 35mm 10mm, clip=true]{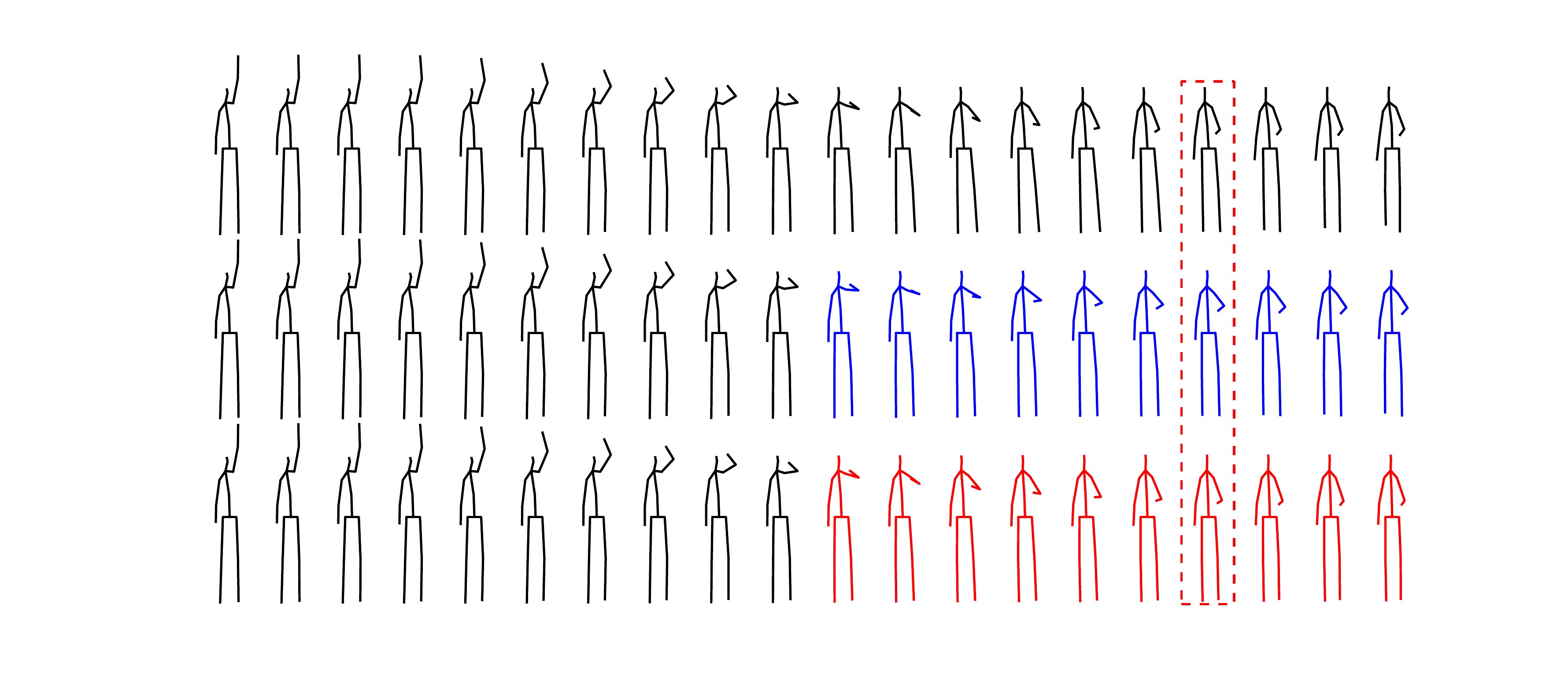} % width=0.9\columnwidth  width=\linewidth
\label{vish36m1}
}
\quad
\subfigure[Taking photo]{\includegraphics[width=0.87\columnwidth,height=1in, trim = 48mm 18mm 35mm 10mm, clip=true]{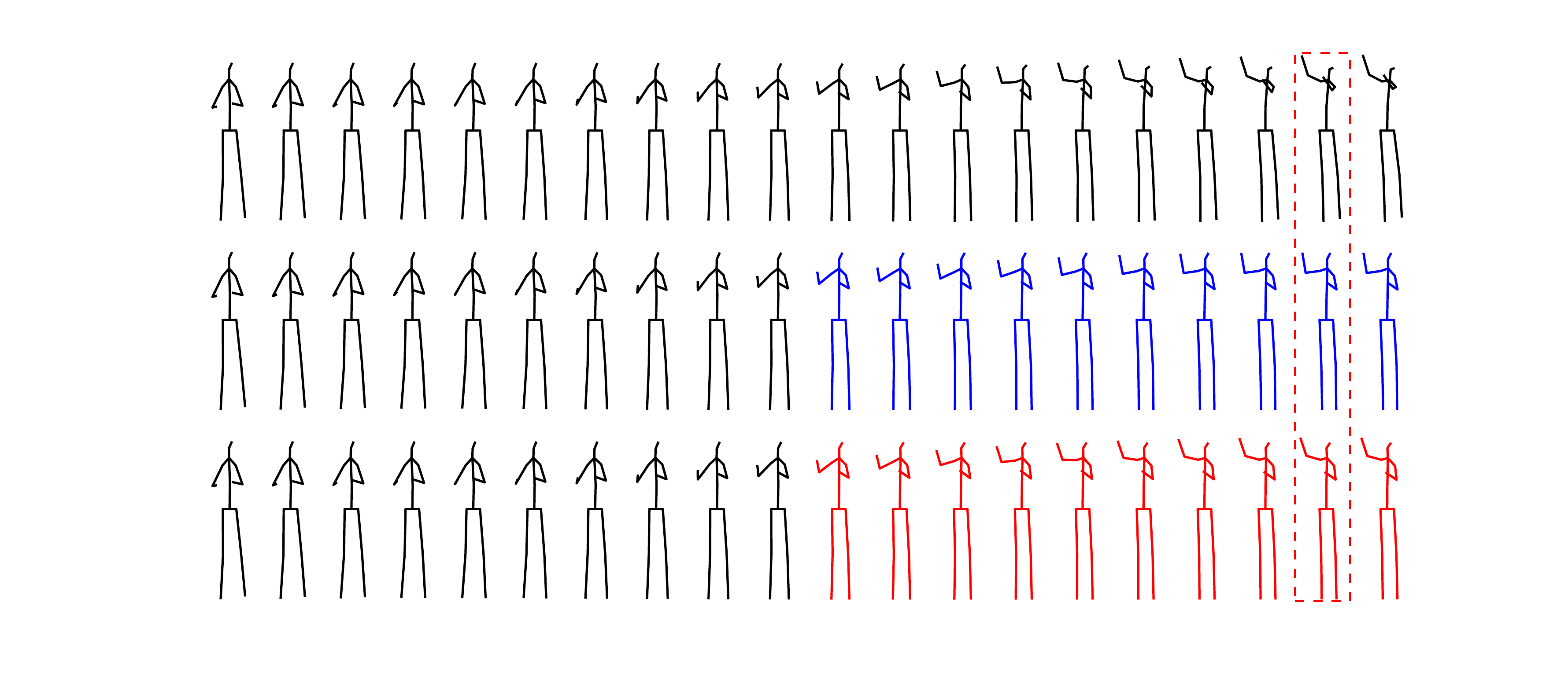}
\label{vish36m2}}
\subfigure[Waiting]{\includegraphics[width=1.8\columnwidth,height=1in,trim = 48mm 15mm 35mm 10mm, clip=true]{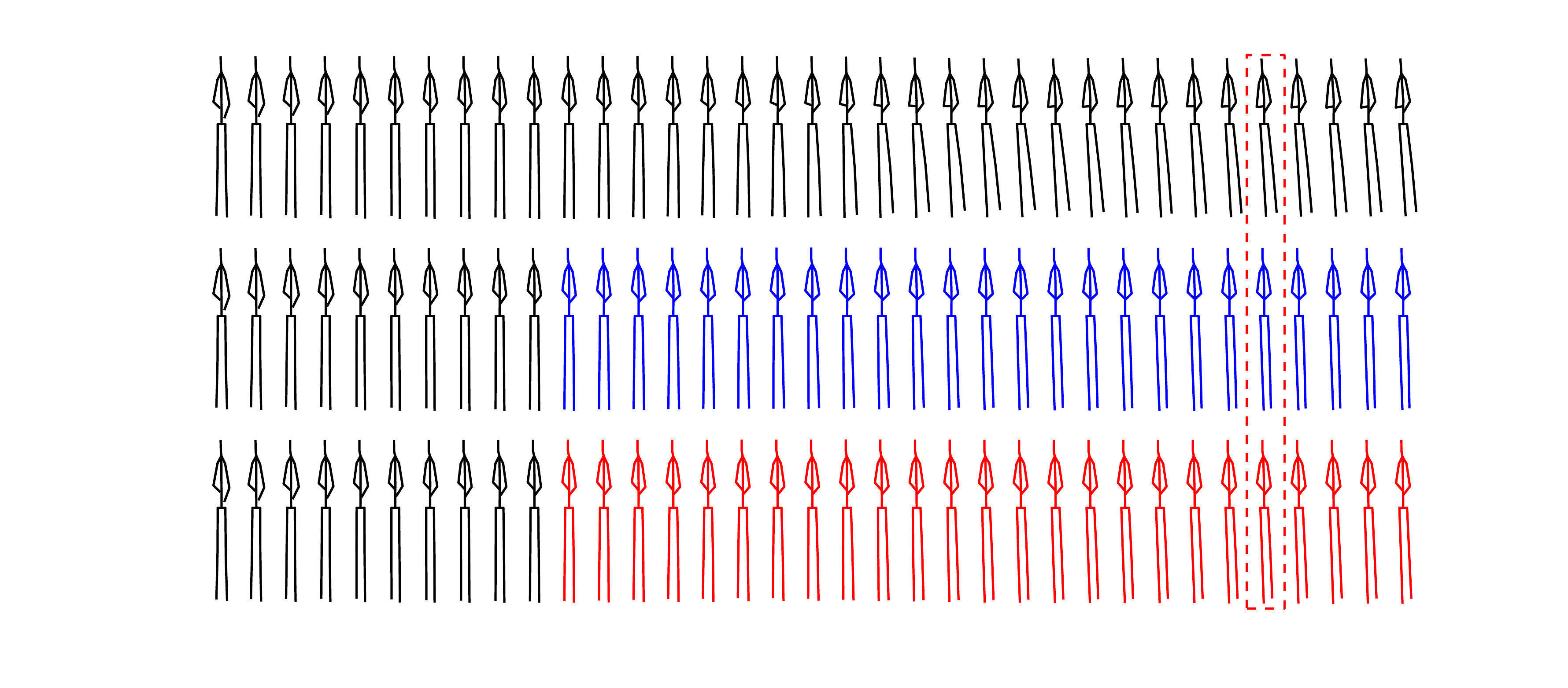}
\label{vish36m3}
}
\end{center}
\caption{Visualization of frame-wise performance on H$3.6$M. (a) and (b) are the results for short-term prediction, (c) is the results for long-term prediction. For each group poses, from top to bottom, we show the groundtruth, the results of LearnTrajDep \cite{TrajD2019}, and the results of our proposed method.}
\label{vish36m}
\end{figure*}

% 可视化图 G3D and FNTU
\begin{figure*}[!t]
\begin{center}
\subfigure[Visualizing results on G$3$D]{\includegraphics[width=1.9\columnwidth,height=1.2in,trim = 48mm 18mm 35mm 75mm, clip=true]{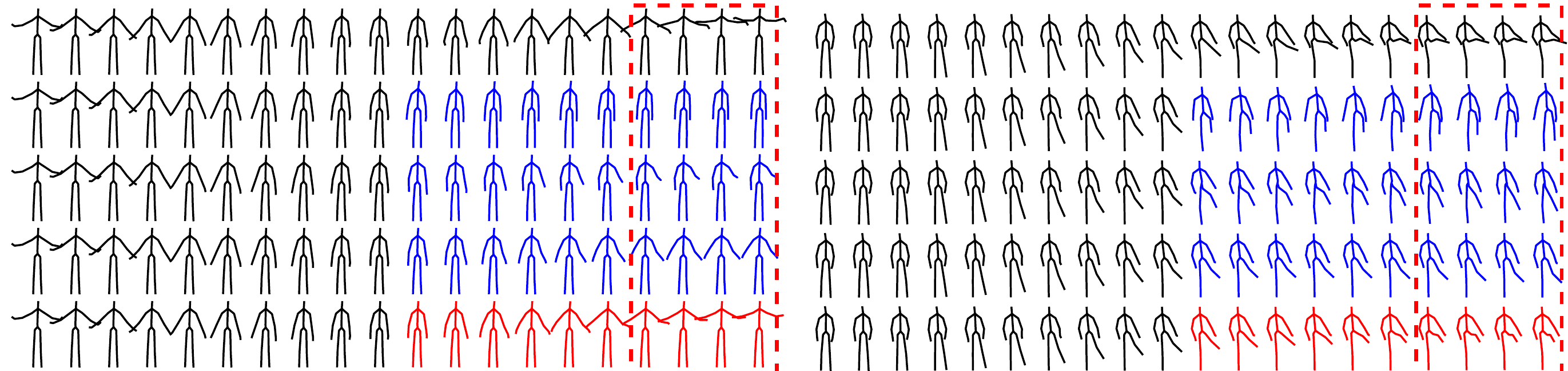}
\label{thirdg}
}
\subfigure[Visualizing results on FNTU]{\includegraphics[width=1.9\columnwidth,height=1.2in,trim = 48mm 18mm 35mm 75mm, clip=true]{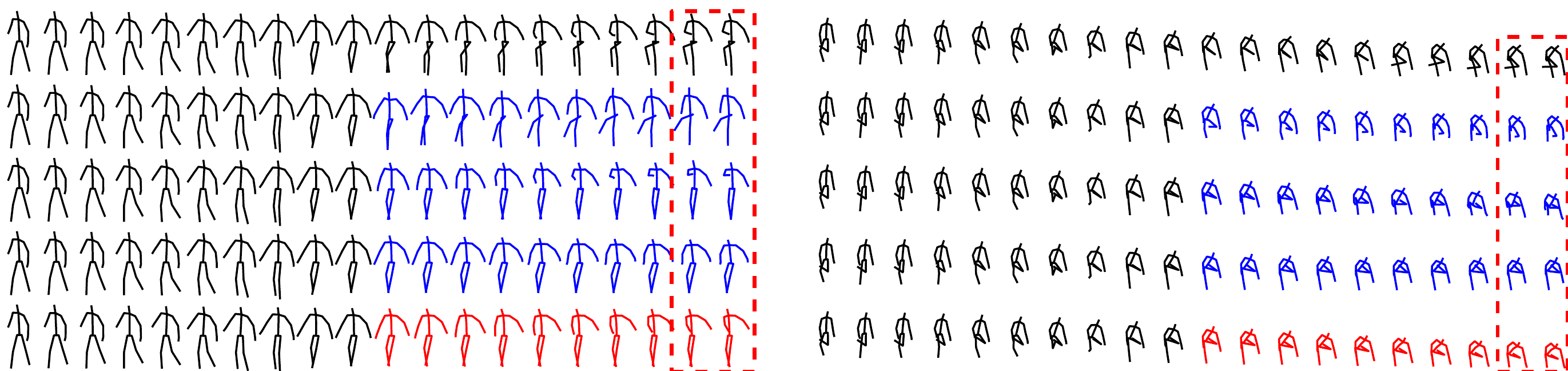}
\label{thirdn}
}
\end{center}
\caption{Visualization of frame-wise performance on G$3$D and FNTU. For each group poses, from top to bottom, we show the groundtruth, the results of PredCNN \cite{predcnn,pisepp}, the results of PISEP$^2$ \cite{pisepp}, the results of LearnTrajDep \cite{TrajD2019} and the results of our proposed method.}
\label{third}
\end{figure*}

To show the qualitative performance of our proposed network, we also visualize the predictive results frame-by-frame on H$3.6$M, G$3$D, and FNTU. As is shown in Fig. \ref{vish36m}, our method achieves the best visualized performance for both short-term and long-term prediction on H$3.6$M.  For example, for the left hand of the human body in Fig. \ref{vish36m1},  the right hand of the human body in Fig. \ref{vish36m2} and the right hand of the human body in Fig. \ref{vish36m3}, compared with the groundtruth, our predictive poses are more reasonable. This may benefit from our proposed network that can end to end capture motion dynamics by simultaneously extracting the local-global spacial features, the coupled spatio-temporal features and the global temporal co-occurrence features of the input human motion sequence, and predict future human motion sequence. So that our proposed model can handle the details of predictive poses better.

Fig. \ref{third} shows the visualization performance on G$3$D and FNTU, and our method achieves the best performance on both datasets. As is denoted in Fig. \ref{thirdg}, compared with \cite{TrajD2019}, two hands of the left group sequence and the left leg of the right group sequence are the closest to the groundtruth.
As is shown in Fig. \ref{thirdn}, compared with \cite{TrajD2019}, for the right hand of the left group sequence and the left leg of the right group sequence,
our predictive poses are more plausible, which demonstrates the effectiveness of our proposed method again.

\section{Conclusion}
In this work, we propose an effective spatio-temporal network, TrajectoryNet, to capture the spatio-temporal dynamics of the previous human motion sequence and predict future human motion sequence in an end to end manner. A major difference between our method and other existing methods is that our model captures the spatio-temporal dynamics of input pose sequence in the trajectory space while other methods model their motion dynamics in the original pose space or the frequency domain.
More importantly, our new proposed network can simultaneously capture the coupled spatio-temporal information and global temporal co-occurrence dependencies of the previous poses, moreover, it can model the different correlations of the joints of different limbs. Evaluations are carried out on three benchmark datasets, and our method achieves state-of-the-art performance on three datasets. Experiments also show that simultaneously modeling the spatial and temporal information is critical to the final performance of the network, and the global temporal co-occurrence modeling can improve the performance of the final network, especially in a shallow network.

% if have a single appendix:
%\appendix[Proof of the Zonklar Equations]
% or
%\appendix  % for no appendix heading
% do not use \section anymore after \appendix, only \section*
% is possibly needed

% use appendices with more than one appendix
% then use \section to start each appendix
% you must declare a \section before using any
% \subsection or using \label (\appendices by itself
% starts a section numbered zero.)
%

% use section* for acknowledgment
\section*{Acknowledgment}
This work was supported partly by the National Natural Science Foundation of China (Grant No. 61673192), and BUPT Excellent Ph.D. Students Foundation (CX2019111). The research in this paper used the NTU RGB+D Action Recognition Dataset made available by the ROSE Lab at the Nanyang Technological University, Singapore.

% Can use something like this to put references on a page
% by themselves when using endfloat and the captionsoff option.
\ifCLASSOPTIONcaptionsoff
  \newpage
\fi

% trigger a \newpage just before the given reference
% number - used to balance the columns on the last page
% adjust value as needed - may need to be readjusted if
% the document is modified later
%\IEEEtriggeratref{8}
% The "triggered" command can be changed if desired:
%\IEEEtriggercmd{\enlargethispage{-5in}}

% references section

% can use a bibliography generated by BibTeX as a .bbl file
% BibTeX documentation can be easily obtained at:
% http://mirror.ctan.org/biblio/bibtex/contrib/doc/
% The IEEEtran BibTeX style support page is at:
% http://www.michaelshell.org/tex/ieeetran/bibtex/
%\bibliographystyle{IEEEtran}
% argument is your BibTeX string definitions and bibliography database(s)
%\bibliography{IEEEabrv,../bib/paper}
%
% <OR> manually copy in the resultant .bbl file
% set second argument of \begin to the number of references
% (used to reserve space for the reference number labels box)

\bibliographystyle{IEEEtran}
\bibliography{TrajectoryNet}
% that's all folks
\end{document}